\documentclass[letterpaper, 10 pt, conference]{ieeeconf}  

\IEEEoverridecommandlockouts                              

\usepackage{graphics} 
\usepackage{epsfig} 
\usepackage{amsmath} 
\usepackage{amssymb}  

\usepackage{biblatex}
\nocite{}

\usepackage{url}
\usepackage[utf8]{inputenc}

\usepackage[margin=0.0cm, skip=1pt, belowskip=0pt]{caption}
\usepackage{subcaption}
\usepackage{hyperref}

\usepackage{lipsum}

\newcommand\blfootnote[1]{%
  \begingroup
  \renewcommand\thefootnote{}\footnote{#1}%
  \addtocounter{footnote}{-1}%
  \endgroup
}

\usepackage{mathtools}

\DeclarePairedDelimiterX{\norm}[1]{\lVert}{\rVert}{#1}

\newcommand{\rcero}{r_{0}}   
\newcommand{\krel}{k_{rel}}

\newcommand{\realNums}{\mathbb{R}}
\newcommand{\slipSmall}{{\scriptscriptstyle SLIP}}

\newcommand{\Pp}{\phi}  
\newcommand{\Qp}{\psi}  
\newcommand{\Entropy}{\mathcal{H}}
\newcommand{\Action}{\mathcal{A}}
\newcommand{\State}{\mathcal{S}}
\newcommand{\LossSymm}{\mathbb{L}_{sym}}

\newcommand{\Q}{Q}
\newcommand{\Exp}{\mathbb{E}}
\newcommand{\Transition}{\mathcal{T}}
\newcommand{\state}{\mathbf{s}}
\newcommand{\action}{\mathbf{a}}

\newcommand{\reward}{r}

\newcommand{\gaitPhase}{\phi_{\slipSmall}}

\newcommand{\MirrorA}{\Psi_{a}}
\newcommand{\MirrorS}{\Psi_{s}}
\newcommand{\q}{\mathbf{q}}
\newcommand{\dq}{\mathbf{\dot{q}}}
\newcommand{\policy}{\pi_\Pp(\action | \state)}
\newcommand{\qFunc}{\Q_{\Qp}\left(\state_{t}, \action_{t}\right)}
\newcommand{\bound}{\mathcal{B}}
\newcommand{\mF}{\mathcal{F}}
\newcommand{\boundConst}{\mathcal{B}_{const}}

\newcommand{\comMomentum}{\mathbf{h}_G}

\newcommand*{\length}[1]{%
    \@tempcnta\z@
    \@for\@tempa:=#1\do{\advance\@tempcnta\@ne}%
    The length of the list #1 is \the\@tempcnta.%
}

\title{\LARGE \bf
An Adaptable Approach to Learn Realistic\\ Legged Locomotion without Examples 
}

     \author{Daniel Ordonez-Apraez$^{1}$, Antonio Agudo$^{1}$, Francesc Moreno-Noguer$^{1}$ and Mario Martin$^{2,3}$ 
     
    \vspace{-10mm}
}

\begin{document}


\twocolumn[{%
	\renewcommand\twocolumn[1][]{#1}%
    \maketitle
    \begin{minipage}{\textwidth}
        \begin{center}
            \begin{tabular}{@{}c@{}}
                \includegraphics[height=3.2cm]{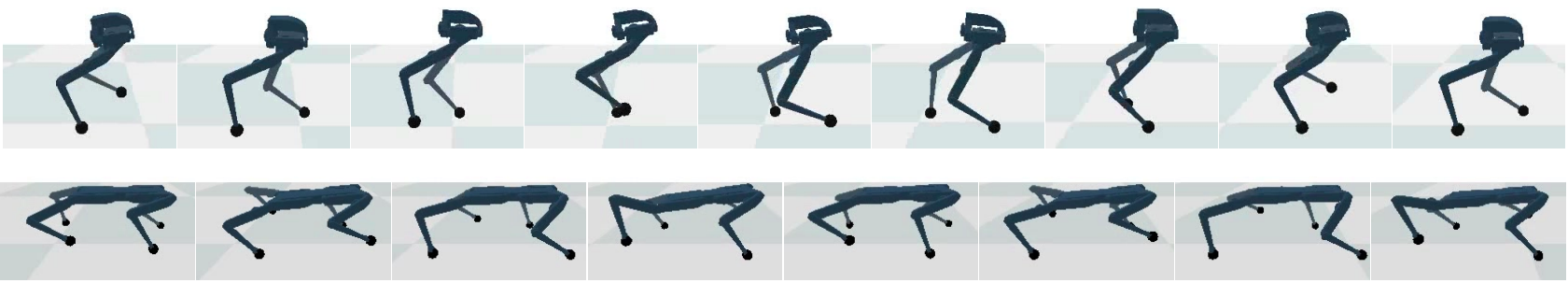}
            \end{tabular}
        \captionof{figure}{Control policies for the biped robot Bolt ($1.05 [m/s]$) and quadruped robot Solo ($0.96 [m/s]$) learned with model-free RL learning guided with the spring-loaded inverted pendulum model. All gaits across training seeds feature periodicity, limbs coordination and natural spring-mass center of mass dynamics.}

        \end{center}    
    \end{minipage}
}]

\blfootnote{\\
$^{1}$ Institut de Rob\`otica i Inform\`atica Industrial, CSIC-UPC, Barcelona, Spain {\tt\footnotesize [dordonez, aagudo, fmoreno]@iri.upc.edu} \\
$^{2}$ Departament de Ciències de la Computació, Universitat Politècnica de Catalunya. $^{3}$ Barcelona Supercomputing Center (BSC).
{\tt\footnotesize mmartin@cs.upc.edu} \\
This work’s experiments were run at the Barcelona Supercomputing Center in collaboration with the HPAI group. This work is supported by the Spanish government with the project MoHuCo PID2020-120049RB-I00 and the ERA-Net Chistera project IPALM PCI2019-103386.}
\begin{abstract}
Learning controllers that reproduce legged locomotion in nature has been a long-time goal in robotics and computer graphics. While yielding promising results, recent approaches are not yet flexible enough to be applicable to legged systems of different morphologies. This is partly because they often rely on precise motion capture references or elaborate learning environments that ensure the naturality of the emergent locomotion gaits but prevent generalization. This work proposes a generic approach for ensuring realism in locomotion by guiding the learning process with the spring-loaded inverted pendulum model as a reference. Leveraging on the exploration capacities of Reinforcement Learning (RL), we learn a control policy that fills in the information gap between the template model and full-body dynamics required to maintain stable and periodic locomotion. The proposed approach can be applied to robots of different sizes and morphologies and adapted to any RL technique and control architecture. 
We present experimental results showing that even in a model-free setup and with a simple reactive control architecture, the learned policies can generate realistic and energy-efficient locomotion gaits for a bipedal and a quadrupedal robot. And most importantly, this is achieved without using motion capture, strong constraints in the dynamics or kinematics of the robot, nor prescribing limb coordination. We provide supplemental videos for qualitative analysis of the naturality of the learned gaits\footnote[4]{\label{footnote:vide_animation} \href{https://youtube.com/playlist?list=PL3ldAwZyOMZHl6Fa28hYOtuGjo8V7Wj6t}{Supplemental videos: https://bit.ly/30NtYAu}}.
\end{abstract}





\section{INTRODUCTION}


The study and generation of natural legged locomotion is a problem that has been approached both by the robotics and computer animation communities. In robotics, the aim is to enable mobile platforms to locomote in various terrains with the flexibility, energy efficiency, and robustness observed in animals in nature. Alternatively, computer graphics focus on generating animations of legged locomotion that appear natural and are compliant to physical laws and interactions with the virtual environment.

Reinforcement Learning, especially in computer graphics, has shown remarkable progress in generating physics-based controllers for complex animated characters in simulation that showcase highly complex and dynamic motion skills \cite{ma2021st_bounds, peng2018deepmimic_, peng2020learning_imitating_dog}. While demonstrating impressive results, the scalability of these methods to robots of different morphologies (i.e., body size, mass, legs anatomy, and number of legs) is somewhat compromised. Mainly because, to ensure realism, they rely on:
    (1) Motion capture references \cite{luo2020carl, peng2020learning_imitating_dog}.
    (2) Precise modeling of the kinematic constraints and approximate dynamics of the musculoskeletal system \cite{jiang2019synthesis_realistic_human,yu2018learning_symmetric_locomotion}.
    (3) Learning environments that feature elaborate curriculum learning techniques \cite{yu2018learning_symmetric_locomotion} or meta-learning of multiple motion skills \cite{hafner2020towards_general_autonomous_learning_of_core_skills_locomotion}.

These approaches aim to aid the learning process by guiding or constraining solution space (i.e., the feasibility region of the optimal policy \cite{ma2021st_bounds}) to avoid converging to unnatural motions. The necessity to do so comes from the fact that for most legged systems, there exist an extensive set of feasible control policies that maintain locomotion; For instance, a human can move forwards at a speed of $2.0 \, [m/s]$ by jogging naturally but also by jumping, limping, using a single leg, and so on. The low variance of gait types in nature across individuals of the same species is a consequence of the tendency of animals to choose the properties of their gaits to maximize stability and robustness while minimizing energy expenditure \cite{alexander2005models_cost_locomotion}.

For most animals in nature, the selection of the optimal gait that minimizes the Cost of Transport (CoT) (i.e., energy used over distance traveled) is a complex and not well-understood function of the forward velocity, terrain properties, and the dynamics of the animal's musculoskeletal system \cite{alexander2005models_cost_locomotion, minetti1997theory_gait_selection}. For this reason, replicating this optimization function in simulation is still an open problem, especially since accurate models of the body dynamics are hard to attain or require considerable design efforts. However, bio-inspired models of legged locomotion are able to approximate valuable properties of the dynamics of locomotion in nature \cite{geyer2005simple_models_compliant, holmes2006dynamics+of_legged_locomotion}, providing substantial queues of the gait properties and gait selection mechanisms in animals.

This work shows that realistic high-speed locomotion can be learned by guiding the control policy with a Spring-Loaded Inverted Pendulum (SLIP) template model. To achieve this, we design an RL environment that employs controlled trajectories of SLIP as reference motions. These are then used to loosely constrain the allowed robot Center of Mass (CoM) trajectories in time, effectively limiting the policy's exploration space to full-body state-action trajectories that feature SLIP CoM  dynamics. 

This learning environment is designed to mimic several properties of full-body CoM trajectory optimization techniques, as the policy is encouraged to learn full-body control that results in energy-efficient CoM trajectories in the vicinity of the reference motion. The main difference is the use of RL to address the optimal control problem instead of traditional convex trajectory optimization methods \cite{bledt2018cheetah, di2018dynamic_cheetah,ding2019realtime_mpc}. With the use of RL we lose the assurances of convex optimization algorithms but gain flexibility in experimentation by avoiding relying on simplifying assumptions of the gait and body dynamics (e.g., centroidal dynamics, zero CoM angular momentum, knowledge of Centers of Pressure (CoP) during contacts, contact timings). Furthermore, our learning setup is compatible with both model-based and model-free RL and with the use of any feed-forward controller. 

Having a generic approach to learning realistic legged locomotion in simulation can be a valuable tool for gait analysis and robot and control design. Some immediate practical applications are the estimation of design requirements (e.g., joint and actuator profiles), identification of passive components, or generation gait information (e.g., centroidal or CoM linear and angular momentum trajectories) to improve the performance of trajectory optimization methods \cite{bledt2020extracting_heuristics_with_reg_MPC}. This work takes a small step towards such a generic system.

\section{BACKGROUND}
\label{sec:background}

    \subsection{Spring Loaded Inverted Pendulum}
    \label{sec:background_slip}
    
\begin{figure}[t] 
         \centering
        \begin{tabular}{@{}cc@{}}
            \includegraphics[height=3.4cm]{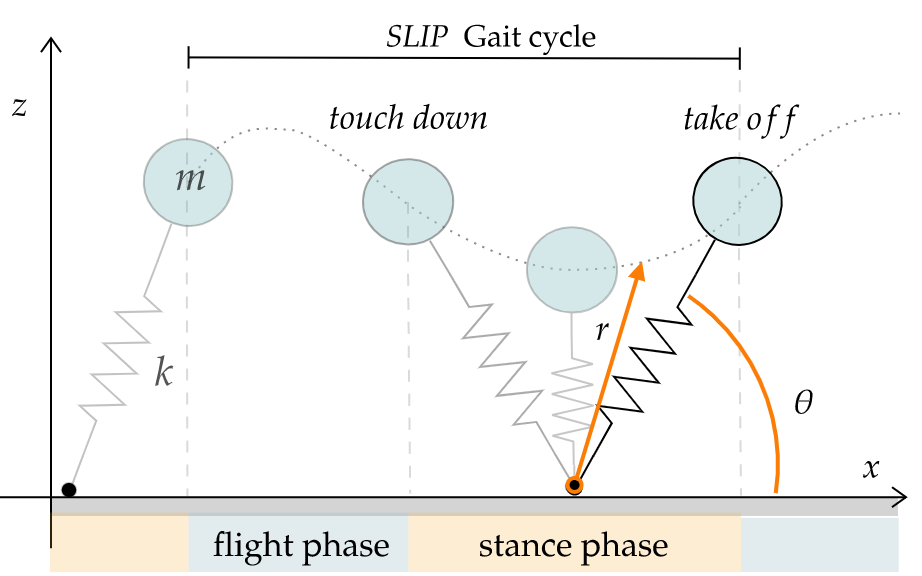}&
            \includegraphics[height=3.4cm]{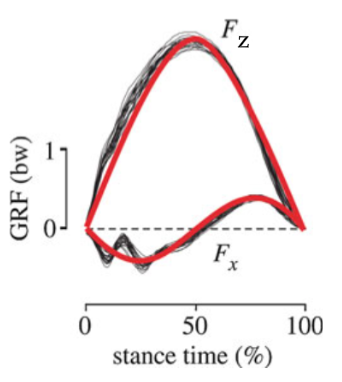}
        \end{tabular}
        \caption{\textbf{SLIP Model parameters and stance cycle.} \textbf{Left:} SLIP gait cycle with flight and stance phases. $m$ and $k$ are the model parameters mass and spring stiffness constant, respectively. \textbf{Right:} SLIP predicted ground reaction forces as a function of the stance time is plotted in red. Traces of ground reaction forces observed in human running experiments in black. Right image adapted from \cite{geyer2018gait}.}
        \label{fig:slip_model}
    \end{figure}

    
    SLIP is a low-order template model for the dynamics of legged locomotion in the sagittal plane. In its simplest form, the model comprises of a point-mass $m$ representing the body's center of mass and a massless spring-leg that models compliance during contact with the ground (see Fig.~\ref{fig:slip_model}-left)~\cite{geyer2018gait}.
    
    Although the model highly simplifies the dynamics of locomotion, it can predict the CoM dynamics' low-frequency components in the sagittal plane (see Fig.~\ref{fig:slip_model}-right)~\cite{geyer2018gait, geyer2005simple_models_compliant, holmes2006dynamics+of_legged_locomotion}. The potential of SLIP as a generic model for animal locomotion in nature was proven by Full et al. \cite{full2000comparative_legged_locomotion} in a comparative study of the dynamics of high-speed locomotion across different animal species and body sizes. The study revealed that most animals experience similar profiles of Ground Reaction Forces (GRFs) while running and that these forces could be approximated with SLIP models. Furthermore, there was evidence of patterns in the parametrization of the model for each animal and gait type, allowing to approximate the animal's GRF during locomotion with a SLIP model parameterized with the animal's mass $m$, hip height $\rcero$, and spring constant $k$ given by:
    \begin{equation}
        k = \frac{k_{rel}\,m\,g}{\rcero}\;\;\;\;  \textrm{with} \;\;\;\; k_{rel}=\frac{\frac{F}{mg}}{\frac{\Delta r}{\rcero}},
        \label{eq:slip_k}
    \end{equation}
    where $F$ is the maximum GRF measured for each animal in the stance phase, $\Delta r$ indicates an approximated maximum vertical displacement of the CoM during the stance phase, $\krel$ represents a dimensionless spring constant used to compare animals across species, and $g$ is the gravity constant. The parametrization of SLIP models' spring constant is usually guided by the distribution of $\krel$ values observed in nature (see further details in \cite{full2000comparative_legged_locomotion}).  
    
    \textit{Information encoded in SLIP dynamics.} The SLIP model dynamics provides insides on the expected time of contact during a stance phase (as a function of the number of legs in contact, the spring constant, and velocity at touch-down), the expected ground reaction forces, and gait period. Additionally, it models the natural fluctuations of the animal's kinetic, potential, and elastic potential (of muscles) energies during locomotion \cite{full2000comparative_legged_locomotion,geyer2018gait}.
    
\subsection{Maximum Entropy Reinforcement Learning}
 Maximum entropy RL is a framework that exploits the concept of entropy from information theory~\cite{carter2007introduction, haarnoja2018soft_actor_critic} and introduces it to the classical RL policy objective to address the known challenge of exploration-exploitation ~\cite{sutton2018reinforcement}. In this framework, agent's policies are modeled by parametric probability distributions, and the agent is tasked with learning to succeed at a given task by maximizing the expected reward while at the same time acting as randomly as possible (i.e., maximize the policy distribution entropy) \cite{haarnoja2018soft_actor_critic}. The augmented policy objective is therefore defined as:
\begin{equation}
          \pi^* = \underset{\pi}{\textrm{argmax}} \sum_{t} \Exp_{(\state_{t}, \action_{t}) \sim \rho_{pi}} \left[ r(\state_{t}, \action_{t})  + \alpha \Entropy(\cdot| \state_{t}) \right], 
    \label{eq:max_entropy_objective}
\end{equation}
where $\pi$ represents the RL agent policy, $\Entropy(\cdot|\state_{t})$ refers to the entropy of the policy given the present state, $\alpha$ the temperature parameter for entropy maximization, and $r(\state_t, \action_t)$ represents the environment reward as a function of the agent present state $\state_t$ and action $\action_t$. 

Maximum entropy RL has been wildly adopted in the robotics community~\cite{haarnoja2018sac_learning_to_walk_dlr, hafner2020towards_general_autonomous_learning_of_core_skills_locomotion,lee2020learning_locomotion_challenging_terrain_eth,peng2020learning_imitating_dog} as its reduced sample complexity is promising for real-life applications. In this work, we use Soft Actor-Critic (SAC) \cite{haarnoja2018soft_actor_critic}, an off-policy actor-critic algorithm that has shown remarkable sample complexity, exploration capacities and requires almost no hyper-parameter tuning. This algorithm was used for real-world locomotion learning with success in \cite{haarnoja2018sac_learning_to_walk_dlr,li2020learning_hierarchical_skills_rl}. 

\subsection{Space-Time Bounds}
\label{sec:background_st_bound}

Space-Time bounds is a technique introduced in \cite{ma2021st_bounds} to ensure that a control policy learns dynamic skills while remaining in the close neighborhood of a reference motion. The technique uses an early-termination criteria to stop the RL episodes whenever the agent's state at a given time $\state_t$ violates a fixed or dynamical state constraint (e.g., detection of robot fall or a velocity limit violation). In our work, the reference SLIP motions are in CoM space rather than in joint space or cartesian/system space (see \cite{orin2008centroidal_momentum_matrix} for details).

By constraining the episode state dynamics with a space-time bound $\bound$, we effectively ensure that the control policy learned through experimentation generates causal state trajectories that remain within $\bound$. This set of possible causal state trajectories allowed by $\bound$ is referred to as the feasible region (or solution space) $\mF(\bound)$ of the control policy. 

\section{RELATED WORK}
\label{sec:related_work}

\subsection{Learning Realistic Legged Locomotion}

Previous locomotion learning approaches use various forms of curriculum learning techniques to guide the learning process and reduce exploitation of simulation inaccuracies or unmodeled dynamics by the learned policy. The benchmark tasks on locomotion train an agent to progressively maximize its forward velocity during learning, using the increasing velocity as a mechanism for pruning unstable and undesired gaits \cite{duan2016benchmarking_rl_continuous_control}. Other approaches rely on modeling more realistically musculoskeletal system dynamics; for instance, in \cite{yu2018learning_symmetric_locomotion}, the authors tune the damping and friction coefficients of the DoF of each robot to encourage realism, and aid the training process with a curriculum learning that stabilizes the robot during the early stages of training. Another work \cite{jiang2019synthesis_realistic_human}, approximates realistic state-dependent joint torque limits of humanoid robots to ensure the emergence of realistic motions using both trajectory optimization techniques and model-free RL.  Lastly, an alternative technique to encourage realism is to train the robots in parallel on multiple motion tasks in a meta-learning environment, reducing the capacity of the policy to exploit unnatural gaits that are functional for single-tasks environments \cite{hafner2020towards_general_autonomous_learning_of_core_skills_locomotion}. 

\begin{center}
    \begin{tabular}{l | c c c c c}
      &     &      & \multicolumn{3}{|c}{Revolute joints limits}  \\     
      & $r_0$[cm] & m[kg] & Pos[rad] & Vel[rad/s] & Torque[N/m] \\ 
      \hline
      Bolt & 35 & 1.3 & $[-\pi, \pi]$ & $[-4\pi, 4\pi]$ & [-2.7, 2.7] \\  
      Solo & 24 & 2.2 & $[-\pi, \pi]$ & $[-4\pi, 4\pi]$ & [-2.7, 2.7] 
    \end{tabular}
     \captionof{table}{Bolt and Solo hip height $\rcero$ and mass $m$ alongside their joints position, velocity and torque limits.}
    \label{table:robot_properties}
\end{center}

\subsection{Guiding controllers with low-order template models} 
Using low-order template models to design or guide controllers is not a novel idea. The assumption of centroidal dynamics is a common technique to enable the convex trajectory optimization methods that fuel most state-of-the-art platforms \cite{bledt2018cheetah,di2018dynamic_cheetah}. These controllers are sensitive to initialization \cite{bledt2020extracting_heuristics_with_reg_MPC}, require prescribed contact timings, and struggle to perform in unstructured environments \cite{lee2020learning_locomotion_challenging_terrain_eth}.

In the context of RL, a bipedal extension of the SLIP model was used to successfully guide the learning process of a walking gait for the Cassie robot~\cite{green2021learning_cassie_slip}. 

\section{METHODOLOGY}
\label{sec:methodology}

In this work, we evaluate the potential use of the dynamics of the SLIP template model in the problem of learning controllers for legged locomotion that achieve natural and realistic behavior. The motivations to study this problem are the following: 
\begin{itemize}
    \item The SLIP model provides a flexible mechanism for approximating the CoM dynamics of high-speed legged locomotion observed in nature, making it an ideal candidate for generating reference motions that bias the learned control policy to appear natural. 
    \item A well-structured RL environment offers the ideal learning flexibility to learn control policies that fill in the information gap between the template model CoM dynamics and the robot full-body dynamics while making no assumptions about the legged system morphology.  
    \item Realistic and energy-efficient gaits of high-speed locomotion learned in simulation can become valuable assets in studying the role of individual legs and CoM momentum during gaits.   
\end{itemize}

Our experiments focus on learning control policies that maintain a constant forward velocity through a periodic, realistic, and energy-efficient gait. We test our approach on the bipedal robot Bolt (6 DoF) and a quadrupedal robot Solo (8 DoF)  \cite{grimminger2020solo_bolt}. The robots' kinematics can resemble those of an Ostrich and Cat, but in reality, they have dynamics that differ significantly from any animal in nature, as they are abnormally lightweight and possess joints that are greatly under-constrained and overpowered (see table~\ref{table:robot_properties}).

For each robot, we independently learn control policies for moving at constant velocities of medium and high magnitudes (i.e., 3 and 6 times the robot's hip height per second). For Bolt we consider $\dot{x}_{des} = [1.05, 2.10]\,[m/s]$ and for Solo $\dot{x}_{des} = [0.60, 0.96]\,[m/s]$.

\subsection{Enforcing SLIP dynamics}
\label{sec:methodology_st_bound_slip}

\begin{figure}[t!]
    \centering
    \includegraphics[width=0.93\linewidth, trim={5mm 0.0mm 5mm 0.0cm}]{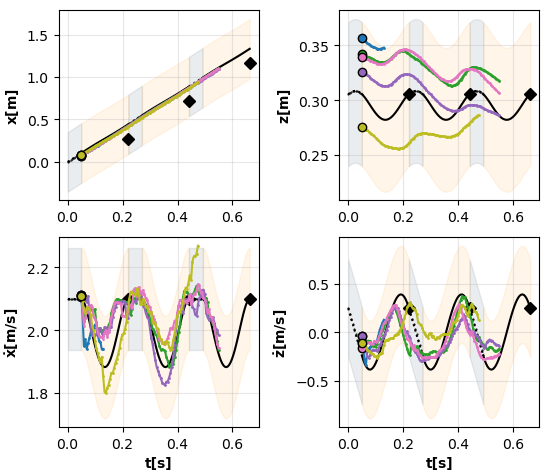}
    \caption{\textbf{Reference SLIP gait CoM trajectory} in the sagittal plane for Bolt maintaining a forward velocity of $2.1 [m/s]$ with an Space-Time Bound $\bound$ with $\epsilon=0.75$ (shaded area). Gray and yellow areas represent flight and stance phases of the SLIP gait, respectively. Experimental trajectories in the early stages of training are plotted to illustrate the boundary violation scenarios and episode terminations.}
    \label{fig:st-bound-experimental}
\end{figure}

The reference motions for each robot and forward velocities are obtained through the optimal control of a SLIP model \cite{chen2019diff_flat_optimal_control_slip,ordonez2021learning_to_run_naturally} that is parameterized with the total robot mass ($m$), the hip height ($\rcero$) \cite{pontzer2007effective_limb_height}, and a relative spring constant $k_{rel}=10.7$ according to \cite{full2000comparative_legged_locomotion, geyer2005simple_models_compliant}. The number of legs expected to contact the ground at the same time during a stance phase is 1 for the bipedal robot Bolt and $2$ for the quadrupedal Solo.

Considering that the SLIP model only approximates the low-frequency CoM dynamics during legged locomotion \cite{full2000comparative_legged_locomotion,holmes2006dynamics+of_legged_locomotion} (see Fig.~\ref{fig:slip_model} right), during learning, we loosely constrain the dynamics of the robot CoM with the reference motion by terminating the learning episode when the robot violates a space-time bound around the reference SLIP CoM trajectory. This condition is evaluated by considering the bound $\bound(t)$ at every time step $t$ in simulation as: 
\begin{align}
         \bound(t) &= \{ \state_{G}(t) \; | \; \rho > |\state_G(t) - \state_\slipSmall(t)| \}, \\
         \state_{G} &= [x_G, y_G, z_G, \dot{x}_G, \dot{y}_G, \dot{z}_G], \\
         \state_{\slipSmall} &= [x_{S}, y_{S}, z_{S}, \dot{x}_S, \dot{y}_S, \dot{z}_S],
    \label{eq:slip_st_bound}
\end{align}
where $\state_{G}$ represents the robot's CoM state in the sagittal plane, $\state_{\slipSmall}$ is the reference SLIP trajectory state, and $\rho$ is a vector that controls the size of the bound (see Fig.~\ref{fig:st-bound-experimental}).

To independently avoid tuning the size of each dimension of the bound for each robot and forward velocity, we parameterize $\rho$ utilizing the robot dimensions and the present gait cycle of the reference trajectory as:   
\begin{equation}
    \begin{split}
    \rho = \epsilon \,
    \left[
        \infty,   
        \frac{\rcero}{2},
        \frac{\rcero}{4},
        \hat{\dot{x}}_S, 
        \frac{\hat{\dot{z}}_S}{2},  
        \frac{\hat{\dot{z}}_S}{2}
    \right]
    \end{split},
    \label{eq:st_bound_size}
\end{equation}
where $\hat{\dot{x}}_S$ and $\hat{\dot{z}}_S$ represent the difference between the maximum and minimum forward and vertical velocities of the SLIP trajectory in the entire present gait cycle, respectively. Lastly, $\epsilon$ is a scalar coefficient that defines the size of the space-time bound. 

\textit{Space/Energy-Time bounds.} The bound $\bound$ in Eq.~\eqref{eq:slip_st_bound} and its implications have a clearer intuition in terms of the system energy: By constraining the robot's CoM height and velocity in the sagittal plane, we effectively bound the robot's potential and CoM kinetic energies in time. Therefore, the imposed feasibility region of the control policy $\mF(\bound)$ constrain the possible locomotion gaits to those that behave energetically similarly to the template model, i.e., gaits that have variations in phase of the potential and CoM kinetic energy. If the bound is sufficiently tight, this encourages the robot to use its legs to mimic the energy storage and release mechanisms observed in legged systems in nature during high-speed locomotion~\cite{geyer2018gait,holmes2006dynamics+of_legged_locomotion}.  

\subsection{Reinforcement Learning Environment}

The Markov Decision Process (MDP) used in this work is defined by the tuple $\langle \State, \Action, \reward, \Transition, \gamma, s_0 \rangle$. $\State$ and $\Action$ are the state and action spaces, respectively. $\reward : \State \times \Action \rightarrow \realNums+$ is a positive reward function defined for every state action pair. $\Transition : \State \times \Action \rightarrow \State$ represents the environment dynamics transition function, governed by the physics simulator Bullet \cite{coumans2021_pybullet}. Moreover, $\gamma$ is the discount factor, and $s_0$ indicates the initial state distribution.

As in \cite{ma2021st_bounds}, the reward is a positive real function so that the policy is encouraged to remain within $\mF(\bound)$. The reward can be represented by $r = r_S \cdot r_P$, where $r_S \in \{0, 1\}$ is a survival reward taking the value of $1$ whenever the robot state is within the bound $\bound$ and $0$ otherwise, and $ r_P \in [0, 1]$ is a multiplicative energy and torque minimization reward, similar to the ones used in \cite{duan2016benchmarking_rl_continuous_control}, defined as: 
\begin{equation}
    r_{P} = \left(1 - \cfrac{\norm{\mathbf{\tau} \dot{q}}}{\norm{\mathbf{\tau}_{max} \dot{q}_{max}}} \right) \cdot \left( 1 - \cfrac{\norm{\mathbf{\tau}}}{\norm{\mathbf{\tau}_{max}}} \right) ,    
\end{equation}
where $\mathbf{\tau}$ is the joint torques and $\dot{\q}$ the joint velocities.

Intuitively, the early episode termination using the bound $\bound$ constrains the exploration of the RL agent to remain close to the reference SLIP trajectory in CoM space, i.e., the agent will only accumulate experiences of causal state-action trajectories that maintain the robot's CoM within the $\mF(\bound)$. At the same time, due to the positive reward, as learning progresses, the agent will try to survive by following state-action trajectories of a longer duration, while the energy minimization reward penalizes inefficient trajectories.  

This learning environment was designed as a CoM full-body trajectory optimization: The RL exploration (in state-action space) is constrained to the local vicinity of the reference trajectory (in CoM space) and the cost/reward function is composed solely of an energy minimization term.  






\subsection{Maximum-Entropy RL and Control architecture}
The learning setup of this work is framed as a Maximum Entropy Inverse Reinforcement Learning (IRL) task~\cite{mai2019irl_maximum_causal_entropy,ziebart2010maximum_entropy_causal}, where the SLIP dynamics are used as the demonstrations and the objective of the optimization is to find the optimal control policy that maximizes the expected reward and policy distribution entropy (see Eq.~\eqref{eq:max_entropy_objective}). The algorithm used is SAC with automatic temperature adjustment~\cite{haarnoja2018soft_actor_critic}.

The policy $\policy$ and Q functions $\qFunc$ of SAC are parameterized with Fully-Connected Neural Networks $\Pp$ and $\Qp$ with two hidden layers of 256 neurons and ReLu activation functions. As in~\cite{haarnoja2018soft_actor_critic}, the control policy is modeled as a state conditional $tanh$ transformed normal distribution $\policy \sim \tanh{\mathcal{N}(\mathbf{\mu}_{\Pp}, \Sigma_{\Pp})}$ with mean and diagonal covariance matrix parameterized by $\Pp$. All experiments are run with $\gamma=0.995$ and a learning rate of $lr=3e^{-4}$. Lastly, the control frequency is set to $200Hz$, considering the reactive nature of the policy.

\subsection{State and Action space}
We use a reactive policy that directly controls the torques of the robot joints $\mathbf{\tau}$, by scaling the action $\action \sim \pi_\Pp(\action|\state)$ by each DoF torque limits (see table~\ref{table:robot_properties}). The MDP state space $\state$ is defined as: 
\begin{equation}
     [\comMomentum, z_G, R, \q, cos(\gaitPhase), sin(\gaitPhase), \dot{x}_{des} - \dot{x}_G, \mathbf{c}],
\end{equation}
where $\comMomentum = A_G(\q)\dq$ is the robot's CoM momentum expressed in the robot base coordinates, computed using the Centroidal Momentum Matrix (CMM) $A_G(\q)$~\cite{orin2008centroidal_momentum_matrix}. $z_G$ is the robot's base height. $R$ is the 6D continuous representation of the robot's base orientation in world coordinates~\cite{zhou2019continuity_rot}. As in \cite{lee2020learning_locomotion_challenging_terrain_eth}, $\gaitPhase$ is a gait phase signal increasing linearly from $0$ to $\pi$ during the flight phase and from $\pi$ to $2\pi$ on the stance phase of the SLIP gait cycle. Lastly, $\dot{x}_{des}$ is the desired forward velocity, and $\mathbf{c}$ is a binary vector indicating contact of the robot's feet with the ground as in~\cite{yu2018learning_symmetric_locomotion}. Note that the state observation is invariant w.r.t the direction of forward motion.

\subsection{Directional and Sagittal Symmetry}

A relevant property of gaits in nature is their symmetry: Animals, including humans, tend to control their body's left and right parts similarly when not injured. In the context of RL, \cite{zinkevich2001symmetry_mdp} showed that when MDPs have symmetry, the optimal policy, Value, and $\Q$ functions are also symmetrical. While in legged locomotion learning~\cite{yu2018learning_symmetric_locomotion} reported enhanced performance and realism of gaits when the symmetry was encouraged in the control policy.

In this work, we extended the policy symmetry loss in~\cite{yu2018learning_symmetric_locomotion} to function for SAC's policies. The main difference is that~\cite{yu2018learning_symmetric_locomotion} uses a Proximal Policy Optimization (PPO) policy which has fixed variance in their action distribution, while in SAC, the variance for the action distribution is also a function of the input state ($\Sigma_{\Pp}$). Therefore our policy symmetry loss is defined as: 
\begin{align}
        \LossSymm = &\norm{\bar{\pi}_{\Pp}(\state) - \MirrorA(\bar{\pi}_{\Pp}(\MirrorS(\state)))} +\nonumber \\  &\norm{\hat{\pi}_{\Pp}(\state) - |\MirrorA(\hat{\pi}_{\Pp}(\MirrorS(\state)))|\,} , 
\end{align}
where $\bar{\pi}_{\Pp} = \Exp[\pi_{\Pp}(\action| \state)]$ represents the mean value of the policy distribution, and $\hat{\pi}_{\Pp} = \Exp[\pi_{\Pp}(\action| \state) - \bar{\pi}_{\Pp}]$ the variance of the action policy distribution. Lastly, $\MirrorS$ and $\MirrorA$ are functions mapping states and actions to their mirrored equivalents.

Furthermore, influenced by~\cite{zinkevich2001symmetry_mdp}, we encouraged symmetry in the Critic \Q-functions by augmenting each batch sampled from the replay buffer with its symmetric equivalent. 

Note that encouraging symmetry in control actions merely requires the control policy to have similar reactions to symmetric states, but not necessarily symmetric gaits trajectories.   

\section{EXPERIMENTAL RESULTS}
\label{sec:results}

We evaluated our approach by learning control policies for both robots with and without enforcing the SLIP dynamics and symmetry. The control policies trained without enforcing the SLIP dynamics use a space-time bound $\boundConst$ where only the CoM forward velocity and lateral position are bounded. The bound for the forward velocity is centered around the desired constant speed $\dot{x}_{des}$, with $\epsilon = 2.0$ to allow for the expected oscillation around the target velocity proper of animal gaits. Therefore the policy feasibility region $\mF(\boundConst)$ contains any locomotion gait that can remain close to the target velocity. In the rest of the document, we refer to the policies trained using the SLIP space-time bound $\bound$ as \textit{with-SLIP}, and \textit{without-SLIP} for the ones using $\boundConst$. 

\begin{figure*}[t!] 
    \begin{subfigure}[b]{0.246\textwidth}
        \centering
        \includegraphics[width=\textwidth, trim={5.8mm 5mm 0mm 5mm}]{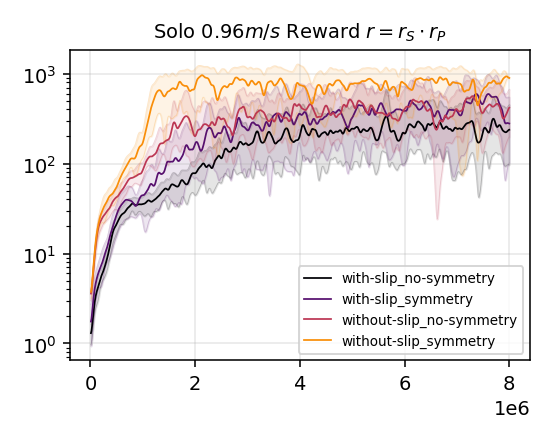}
        \caption{}
    \end{subfigure}
    \begin{subfigure}[b]{0.246\textwidth}
        \centering
        \includegraphics[width=\textwidth, trim={5.8mm 5mm 0mm 5mm}]{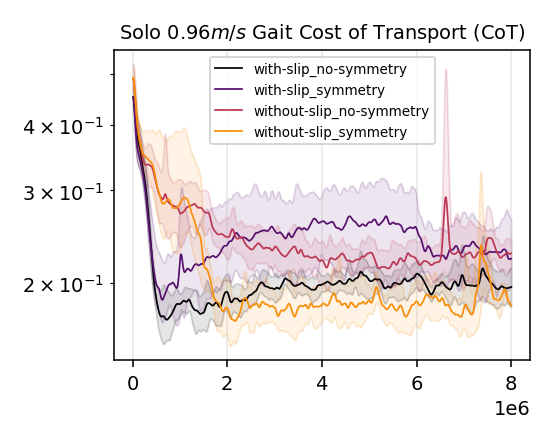}
        \caption{}
    \end{subfigure}
    \begin{subfigure}[b]{0.246\textwidth}
        \centering
        \includegraphics[width=\textwidth, trim={5.8mm 5mm 0mm 5mm}]{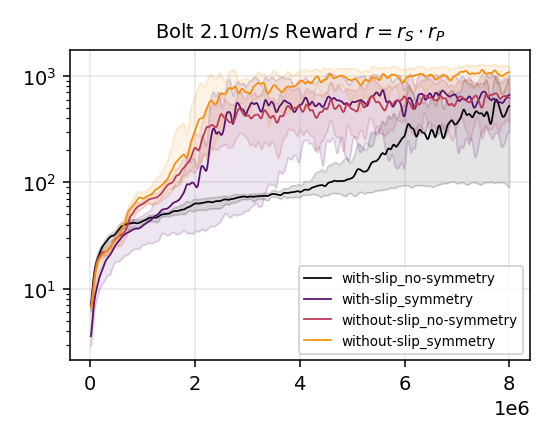}
        \caption{}
    \end{subfigure}
    \begin{subfigure}[b]{0.246\textwidth}
        \centering
        \includegraphics[width=\textwidth, trim={5.8mm 5mm 0mm 5mm}]{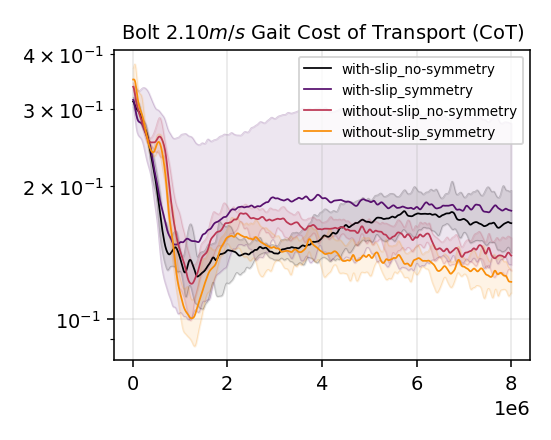}
        \caption{}
    \end{subfigure}
    \caption{\textbf{Reward curves and gait's Cost of Transport} for high speed locomotion of Solo $0.96[m/s]$ (a and b) and for Bolt $2.1[m/s]$ (c and d). All training curves show the step average, maximum and minimum (shaded region) across 5 training seeds. Exploiting symmetry results in more robust control policies (i.e., higher and faster survival rates) across all training variants. For Bolt exploiting symmetry reduces considerable the sample-complexity. Although quantitatively the variants \textit{with-SLIP} and symmetry appear not to perform better, we highly encourage the reader to compare the gait properties qualitatively in the supplemental video \ref{footnote:with_without_video}.}
    \label{fig:quantitative_results}
\end{figure*}

\subsection{Qualitative and quantitative effects of imposing SLIP dynamics}

The control policies \textit{without-SLIP} for both robots find stable gait trajectories that maintain forward velocity within the imposed bound $\boundConst$. However, realistic or natural behavior is not present in any of the training seeds (see comparison video \footnote[4]{\label{footnote:with_without_video} Videos \textit{with-SLIP} vs. \textit{without-SLIP}:  \href{https://youtu.be/\_0Y3sJ9w9F8}{youtu.be/\_0Y3sJ9w9F8}}). Due to the large feasibility region, each training seed rapidly converges to unnatural aperiodic gaits that exploit the unmodeled dynamics (e.g., actuator dynamics, joint-space friction, and damping) and the simulation inaccuracies in contact dynamics (specifically the transition from static to dynamic friction regimes). Furthermore, because the elastic properties of muscles are not modeled, the energy minimization reward $r_P$ leads both robots to locomote with short and rapid stepping with nearly fully extended legs (since such a gait minimizes the required energy and torques for compensating gravity). 

On the contrary, policies trained \textit{with-SLIP} showcase natural limb coordination and spring-mass dynamics, resulting in realistic and periodic gaits. Because there is no penalization on the torso's linear and angular momentum, the robots learn to locomote with graceful and natural torso motions that sync with leg contacts. Both robots perform irregular footstep sequences when needed to stabilize but tend to converge to near-periodic footstep sequences once they achieve to remain in a limit cycle. In all experiments, Solo converges to use a trot gait, coordinating diagonally opposed legs, while Bolt uses a running gait.   

From a quantitative perspective, the control policies without-SLIP showcase similar or better performance in sample complexity, gait robustness, and energetic cost-of transport (see Fig.~\ref{fig:quantitative_results}). Therefore, for the RL agent, the unnatural gaits represent suitable and close to optimal control policies. This minor difference in quantitative performance highlights the challenges of obtaining realistic behavior with learning techniques without introducing inductive biases in the learning process to encourage realism.   

\textit{Gait variance across seeds.} Perhaps the most relevant impact of enforcing SLIP gait dynamics through space-time bounds is the reduction of the variance of gait properties across training seeds, meaning that final policies for each seed converges to locomotion gaits that feature similar limb coordination and CoM trajectories (see all seed's gaits in supplemental video \ref{footnote:with_without_video}). 

\begin{figure}[!t]
    \centering
    \includegraphics[width=0.96\linewidth, trim={5mm 0.0mm 5mm 0.0cm}]{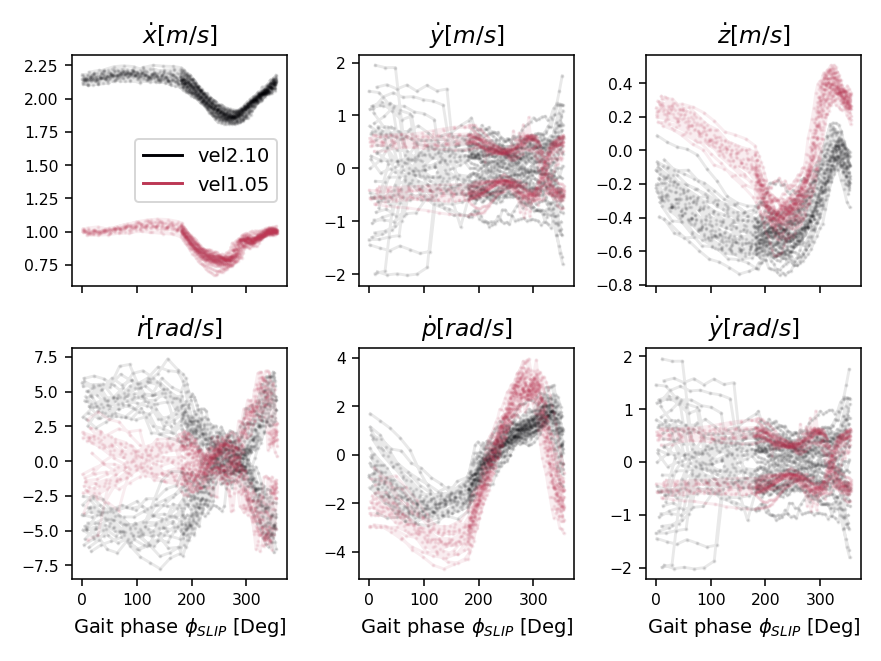}
    \caption{{\bf \textit{With-SLIP} and symmetry linear and angular gait centroidal velocities} of the torso of Bolt, at medium and high forward velocities. The gaits showcase periodicity and spring-mass dynamics proper of the SLIP model. For the lateral velocity $\dot{y}$, roll $\dot{r}$ and yaw $\dot{\mathrm{y}}$ the two trajectory modes correspond to gait cycles with left and right leg contact}
    \label{fig:centroidal_traj}
\end{figure}

\textit{Utility of learned gaits.} Figure~\ref{fig:centroidal_traj} presents the robot base centroidal trajectory during the SLIP gait cycle for the Bolt. This trajectory and those of all training seeds alongside other gait properties can be exploited by centroidal or CoM trajectory optimization methods ~\cite{ding2019realtime_mpc,orin2008centroidal_momentum_matrix,ponton2018time_com_optimization,xie2021glide}.
\subsection{Impact on symmetry enforcing}
The training results in Fig.~\ref{fig:quantitative_results} show a considerable increase in performance by exploiting the symmetry of the robot control and the MDP for both \textit{with-SLIP} and \textit{without-SLIP} variants, as there are signs of improved sample complexity and control robustness (i.e., ability to maintain stable gaits). The impact of symmetry in the realism of the gaits is best observed qualitatively in the comparison video \footnote[5]{\label{footnote:symmetry_video} Videos \textit{with-symmetry} vs. \textit{without-symmetry}:  \href{https://youtu.be/qN25-XQBqKM}{youtu.be/qN25-XQBqKM}}.

\section{CONCLUSIONS AND FUTURE WORK}
\label{sec:conclusions}

In this work, we show the potential use of the SLIP model as a mechanism for measuring and asserting realism in gaits of high-speed locomotion. Taking a step towards locomotion learning without MoCap references. To achieve this, we designed an RL environment capable of producing realistic full-body periodic locomotion gaits for any robot morphology, using a simple energy minimization reward. We present results with a bipedal and quadrupedal robot in simulation for the model-free control case and a reactive policy acting directly on joint torques.

The advantage of our approach relies on the use of controlled trajectories of the SLIP model as a reference to loosely constrain the policy exploration space. This model can be parameterized to approximate the dynamics of high-speed locomotion of most animals/robots. Enabling applications where no motion capture data is available. 

In future work we want to: (1) Enforce symmetry as a hard constraint in the optimization process by using equivariant function approximators for the actor and critic functions \cite{finzi2021emlp}. (2) Expand the experiments scope to embrace a larger suite of robot morphologies (and sizes), terrain properties, and energy penalization schemes (based on various models of the cost of transport \cite{alexander2005models_cost_locomotion}). (3) Explore the use of extended versions of the SLIP model to address the realism of walking gaits of different robot morphologies. (4) Adapt our approach for Computational Design: By studying the learned robot dynamics at different speeds, we theorize, we can test potential passive components for each DoF to optimize gait energy expenditure, control robustness, and/or the robot's kinematics and dynamics.

\printbibliography

@article{yu2018learning_symmetric_locomotion,
  title={Learning symmetric and low-energy locomotion},
  author={Yu, Wenhao and Turk, Greg and Liu, C Karen},
  journal={ACM Transactions on Graphics (TOG)},
  volume={37},
  number={4},
  pages={1--12},
  year={2018},
  publisher={ACM New York, NY, USA}
}

@inproceedings{zinkevich2001symmetry_mdp,
  title={Symmetry in Markov decision processes and its implications for single agent and multi agent learning},
  author={Zinkevich, Martin and Balch, Tucker},
  booktitle={In Proceedings of the 18th International Conference on Machine Learning},
  year={2001},
  organization={Citeseer}
}

@mastersthesis{ordonez2021learning_to_run_naturally,
  title={Learning to run naturally: guiding policies with the Spring-Loaded Inverted Pendulum},
  author={Ordo{\~n}ez Apraez, Daniel Felipe},
  year={2021},
  school={Universitat Polit{\`e}cnica de Catalunya}
}

@article{minetti1997theory_gait_selection,
  title={A theory of metabolic costs for bipedal gaits},
  author={Minetti, AE and Alexander, R McN},
  journal={Journal of Theoretical Biology},
  volume={186},
  number={4},
  pages={467--476},
  year={1997},
  publisher={Elsevier}
}

@article{holmes2006dynamics+of_legged_locomotion,
  title={The dynamics of legged locomotion: Models, analyses, and challenges},
  author={Holmes, Philip and Full, Robert J and Koditschek, Dan and Guckenheimer, John},
  journal={SIAM review},
  volume={48},
  number={2},
  pages={207--304},
  year={2006},
  publisher={SIAM}
}

@inproceedings{ma2021st_bounds,
  title={Learning and Exploring Motor Skills with Spacetime Bounds},
  author={Ma, Li-Ke and Yang, Zeshi and Tong, Xin and Guo, Baining and Yin, KangKang},
  booktitle={Computer Graphics Forum},
  volume={40},
  number={2},
  pages={251--263},
  year={2021},
  organization={Wiley Online Library}
}

@article{grimminger2020solo_bolt,
title={An Open Torque-Controlled Modular Robot Architecture for Legged Locomotion Research},
author={F. {Grimminger} and A. {Meduri} and M. {Khadiv} and J. {Viereck} and M. {Wüthrich} and M. {Naveau} and V. {Berenz} and S. {Heim} and F. {Widmaier} and T. {Flayols} and J. {Fiene} and A. {Badri-Spröwitz} and L. {Righetti}},
journal={IEEE Robotics and Automation Letters},
year={2020},
volume={5},
number={2},
pages={3650-3657},
% doi={10.1109/LRA.2020.2976639},
arXiv = {arXiv:1910.00093}}

@article{pontzer2007effective_limb_height,
  title={Effective limb length and the scaling of locomotor cost in terrestrial animals},
  author={Pontzer, Herman},
  journal={Journal of Experimental Biology},
  volume={210},
  number={10},
  pages={1752--1761},
  year={2007},
  publisher={Company of Biologists}
}

@article{chen2019diff_flat_optimal_control_slip,
  title={Optimal control of a differentially flat two-dimensional spring-loaded inverted pendulum model},
  author={Chen, Hua and Wensing, Patrick M and Zhang, Wei},
  journal={IEEE Robotics and Automation Letters},
  volume={5},
  number={2},
  pages={307--314},
  year={2019},
  publisher={IEEE}
}

@article{green2021learning_cassie_slip,
  title={Learning Spring Mass Locomotion: Guiding Policies with a Reduced-Order Model},
  author={Green, Kevin and Godse, Yesh and Dao, Jeremy and Hatton, Ross L and Fern, Alan and Hurst, Jonathan},
  journal={IEEE Robotics and Automation Letters},
  volume={6},
  number={2},
  pages={3926--3932},
  year={2021},
  publisher={IEEE}
}

@MISC{coumans2021_pybullet,
author =   {Erwin Coumans and Yunfei Bai},
title =    {PyBullet, a Python module for physics simulation for games, robotics and machine learning},
howpublished = {\url{http://pybullet.org}},
year = {2016--2021}
}

@inproceedings{bledt2020extracting_heuristics_with_reg_MPC,
  title={Extracting legged locomotion heuristics with regularized predictive control},
  author={Bledt, Gerardo and Kim, Sangbae},
  booktitle={2020 IEEE International Conference on Robotics and Automation (ICRA)},
  pages={406--412},
  year={2020},
  organization={IEEE}
}

@inproceedings{zhou2019continuity_rot,
  title={On the continuity of rotation representations in neural networks},
  author={Zhou, Yi and Barnes, Connelly and Lu, Jingwan and Yang, Jimei and Li, Hao},
  booktitle={Proceedings of the IEEE/CVF Conference on Computer Vision and Pattern Recognition},
  pages={5745--5753},
  year={2019}
}

@incollection{full2000comparative_legged_locomotion,
  title={Musculoskeletal dynamics in rhythmic systems: a comparative approach to legged locomotion},
  author={Full, Robert J and Farley, Claire T and Winters, Jack M},
  booktitle={Biomechanics and neural control of posture and movement},
  pages={192--205},
  year={2000},
  publisher={Springer}
}

@article{geyer2018gait,
  title={Gait based on the spring-loaded inverted pendulum},
  author={Geyer, H and Saranli, U and Goswami, A and Vadakkepat, P},
  journal={Humanoid Robotics: A Reference, A. Goswami and P. Vadakkepat, Eds. Springer Netherlands},
  pages={1--25},
  year={2018}
}

@phdthesis{geyer2005simple_models_compliant,
  title={Simple models of legged locomotion based on compliant limb behavior | Grundmodelle pedaler Lokomotion basierend auf nachgiebigem Beinverhalten},
  author={Geyer, Hartmut},
  year={2005}
}

@article{hafner2020towards_general_autonomous_learning_of_core_skills_locomotion,
  title={Towards general and autonomous learning of core skills: A case study in locomotion},
  author={Hafner, Roland and Hertweck, Tim and Kl{\"o}ppner, Philipp and Bloesch, Michael and Neunert, Michael and Wulfmeier, Markus and Tunyasuvunakool, Saran and Heess, Nicolas and Riedmiller, Martin},
  journal={arXiv preprint arXiv:2008.12228},
  year={2020}
}

@book{ziebart2010maximum_entropy_causal,
  title={Modeling purposeful adaptive behavior with the principle of maximum causal entropy},
  author={Ziebart, Brian D},
  year={2010},
  publisher={Carnegie Mellon University}
}

@article{mai2019irl_maximum_causal_entropy,
  title={Generalized Maximum Causal Entropy for Inverse Reinforcement Learning},
  author={Mai, Tien and Chan, Kennard and Jaillet, Patrick},
  journal={arXiv preprint arXiv:1911.06928},
  year={2019}
}

@article{haarnoja2018soft_actor_critic,
  title={Soft actor-critic algorithms and applications},
  author={Haarnoja, Tuomas and Zhou, Aurick and Hartikainen, Kristian and Tucker, George and Ha, Sehoon and Tan, Jie and Kumar, Vikash and Zhu, Henry and Gupta, Abhishek and Abbeel, Pieter and others},
  journal={arXiv preprint arXiv:1812.05905},
  year={2018}
}

@article{haarnoja2018sac_learning_to_walk_dlr,
  title={Learning to walk via deep reinforcement learning},
  author={Haarnoja, Tuomas and Ha, Sehoon and Zhou, Aurick and Tan, Jie and Tucker, George and Levine, Sergey},
  journal={arXiv preprint arXiv:1812.11103},
  year={2018}
}

@inproceedings{li2020learning_hierarchical_skills_rl,
  title={Learning generalizable locomotion skills with hierarchical reinforcement learning},
  author={Li, Tianyu and Lambert, Nathan and Calandra, Roberto and Meier, Franziska and Rai, Akshara},
  booktitle={2020 IEEE International Conference on Robotics and Automation (ICRA)},
  pages={413--419},
  year={2020},
  organization={IEEE}
}

@inproceedings{orin2008centroidal_momentum_matrix,
  title={Centroidal momentum matrix of a humanoid robot: Structure and properties},
  author={Orin, David E and Goswami, Ambarish},
  booktitle={2008 IEEE/RSJ International Conference on Intelligent Robots and Systems},
  pages={653--659},
  year={2008},
  organization={IEEE}
}

@article{peng2020learning_imitating_dog,
  title={Learning agile robotic locomotion skills by imitating animals},
  author={Peng, Xue Bin and Coumans, Erwin and Zhang, Tingnan and Lee, Tsang-Wei and Tan, Jie and Levine, Sergey},
  journal={arXiv preprint arXiv:2004.00784},
  year={2020}
}

@article{luo2020carl,
  title={Carl: Controllable agent with reinforcement learning for quadruped locomotion},
  author={Luo, Ying-Sheng and Soeseno, Jonathan Hans and Chen, Trista Pei-Chun and Chen, Wei-Chao},
  journal={ACM Transactions on Graphics (TOG)},
  volume={39},
  number={4},
  pages={38--1},
  year={2020},
  publisher={ACM New York, NY, USA}
}

@article{peng2018deepmimic_,
  title={Deepmimic: Example-guided deep reinforcement learning of physics-based character skills},
  author={Peng, Xue Bin and Abbeel, Pieter and Levine, Sergey and van de Panne, Michiel},
  journal={ACM Transactions on Graphics (TOG)},
  volume={37},
  number={4},
  pages={1--14},
  year={2018},
  publisher={ACM New York, NY, USA}
}

@article{lee2020learning_locomotion_challenging_terrain_eth,
  title={Learning quadrupedal locomotion over challenging terrain},
  author={Lee, Joonho and Hwangbo, Jemin and Wellhausen, Lorenz and Koltun, Vladlen and Hutter, Marco},
  journal={Science robotics},
  volume={5},
  number={47},
  year={2020},
  publisher={Science Robotics}
}

@inproceedings{bledt2018cheetah,
  title={MIT Cheetah 3: Design and control of a robust, dynamic quadruped robot},
  author={Bledt, Gerardo and Powell, Matthew J and Katz, Benjamin and Di Carlo, Jared and Wensing, Patrick M and Kim, Sangbae},
  booktitle={2018 IEEE/RSJ International Conference on Intelligent Robots and Systems (IROS)},
  pages={2245--2252},
  year={2018},
  organization={IEEE}
}

@inproceedings{di2018dynamic_cheetah,
  title={Dynamic locomotion in the mit cheetah 3 through convex model-predictive control},
  author={Di Carlo, Jared and Wensing, Patrick M and Katz, Benjamin and Bledt, Gerardo and Kim, Sangbae},
  booktitle={2018 IEEE/RSJ International Conference on Intelligent Robots and Systems (IROS)},
  pages={1--9},
  year={2018},
  organization={IEEE}
}

@article{xie2021glide,
  title={GLiDE: Generalizable Quadrupedal Locomotion in Diverse Environments with a Centroidal Model},
  author={Xie, Zhaoming and Da, Xingye and Babich, Buck and Garg, Animesh and van de Panne, Michiel},
  journal={arXiv preprint arXiv:2104.09771},
  year={2021}
}

@article{jiang2019synthesis_realistic_human,
  title={Synthesis of biologically realistic human motion using joint torque actuation},
  author={Jiang, Yifeng and Van Wouwe, Tom and De Groote, Friedl and Liu, C Karen},
  journal={ACM Transactions On Graphics (TOG)},
  volume={38},
  number={4},
  pages={1--12},
  year={2019},
  publisher={ACM New York, NY, USA}
}

@inproceedings{ding2019realtime_mpc,
  title={Real-time model predictive control for versatile dynamic motions in quadrupedal robots},
  author={Ding, Yanran and Pandala, Abhishek and Park, Hae-Won},
  booktitle={2019 International Conference on Robotics and Automation (ICRA)},
  pages={8484--8490},
  year={2019},
  organization={IEEE}
}

@book{sutton2018reinforcement,
  title={Reinforcement learning: An introduction},
  author={Sutton, Richard S and Barto, Andrew G},
  year={2018},
  publisher={MIT press}
}

@article{alexander2005models_cost_locomotion,
  title={Models and the scaling of energy costs for locomotion},
  author={Alexander, R McNeill},
  journal={Journal of Experimental Biology},
  volume={208},
  number={9},
  pages={1645--1652},
  year={2005},
  publisher={Company of Biologists}
}

@article{carter2007introduction,
  title={An introduction to information theory and entropy},
  author={Carter, Tom},
  journal={Complex systems summer school, Santa Fe},
  year={2007},
  publisher={Citeseer}
}

@inproceedings{ponton2018time_com_optimization,
  title={On time optimization of centroidal momentum dynamics},
  author={Ponton, Brahayam and Herzog, Alexander and Del Prete, Andrea and Schaal, Stefan and Righetti, Ludovic},
  booktitle={2018 IEEE International Conference on Robotics and Automation (ICRA)},
  pages={5776--5782},
  year={2018},
  organization={IEEE}
}

@inproceedings{duan2016benchmarking_rl_continuous_control,
  title={Benchmarking deep reinforcement learning for continuous control},
  author={Duan, Yan and Chen, Xi and Houthooft, Rein and Schulman, John and Abbeel, Pieter},
  booktitle={International conference on machine learning},
  pages={1329--1338},
  year={2016},
  organization={PMLR}
}

@inproceedings{finzi2021emlp,
  title={A practical method for constructing equivariant multilayer perceptrons for arbitrary matrix groups},
  author={Finzi, Marc and Welling, Max and Wilson, Andrew Gordon},
  booktitle={International Conference on Machine Learning},
  pages={3318--3328},
  year={2021},
  organization={PMLR}
}

\end{document}